\documentclass[letterpaper, 10 pt, conference]{ieeeconf}

\IEEEoverridecommandlockouts                  
\usepackage{graphicx}
\usepackage{bm}
\usepackage[utf8]{inputenc}
\usepackage{makecell}
\usepackage{booktabs}
\usepackage{amsmath}
\usepackage{amsfonts}
\usepackage{acronym}
\usepackage[misc]{ifsym}
\usepackage{float}
\usepackage{amssymb}
\usepackage{graphicx}
\usepackage[colorlinks=true, linkcolor=blue, citecolor=blue]{hyperref}
\usepackage{afterpage}   
\usepackage{stfloats}     
\usepackage[absolute,overlay]{textpos}
\usepackage{tabularx}
\usepackage{ragged2e}
\usepackage{siunitx}
\usepackage[capitalise]{cleveref}
\usepackage{xspace}
\usepackage{array}

\usepackage{xcolor}

\usepackage[table]{xcolor} 
\usepackage{booktabs}      
\renewcommand{\arraystretch}{1.3} 

\crefname{figure}{Fig.}{Figs.}
\Crefname{figure}{Fig.}{Figs.}

\newcommand{\method}[0]{\texttt{SafeFall}\xspace}

\title{\LARGE \bf
\method: Learning Protective Control for Humanoid Robots
}

\author{
Ziyu Meng$^{1,2,*}$,
Tengyu Liu$^{2,*}$,
Le Ma$^2$,
Yingying Wu$^{2,3}$,
Ran Song$^{1,~\textrm{\Letter}}$,
Wei Zhang$^1$,
Siyuan Huang$^{2,~\textrm{\Letter}}$
\vspace{6pt}\\
    \small $^1$ School of Control Science and Engineering, Shandong University
    $^2$ National Key Laboratory of General Artificial Intelligence, BIGAI\\
    \small $^3$ Department of Automation, Tsinghua University
    \vspace{6pt}\\
    \href{https://safefall.github.io}{https://safefall.github.io}
}

\usepackage{cuted}
\usepackage{capt-of}

\begin{document}

\maketitle

\begin{strip}
\centering
\includegraphics[width=1\textwidth]{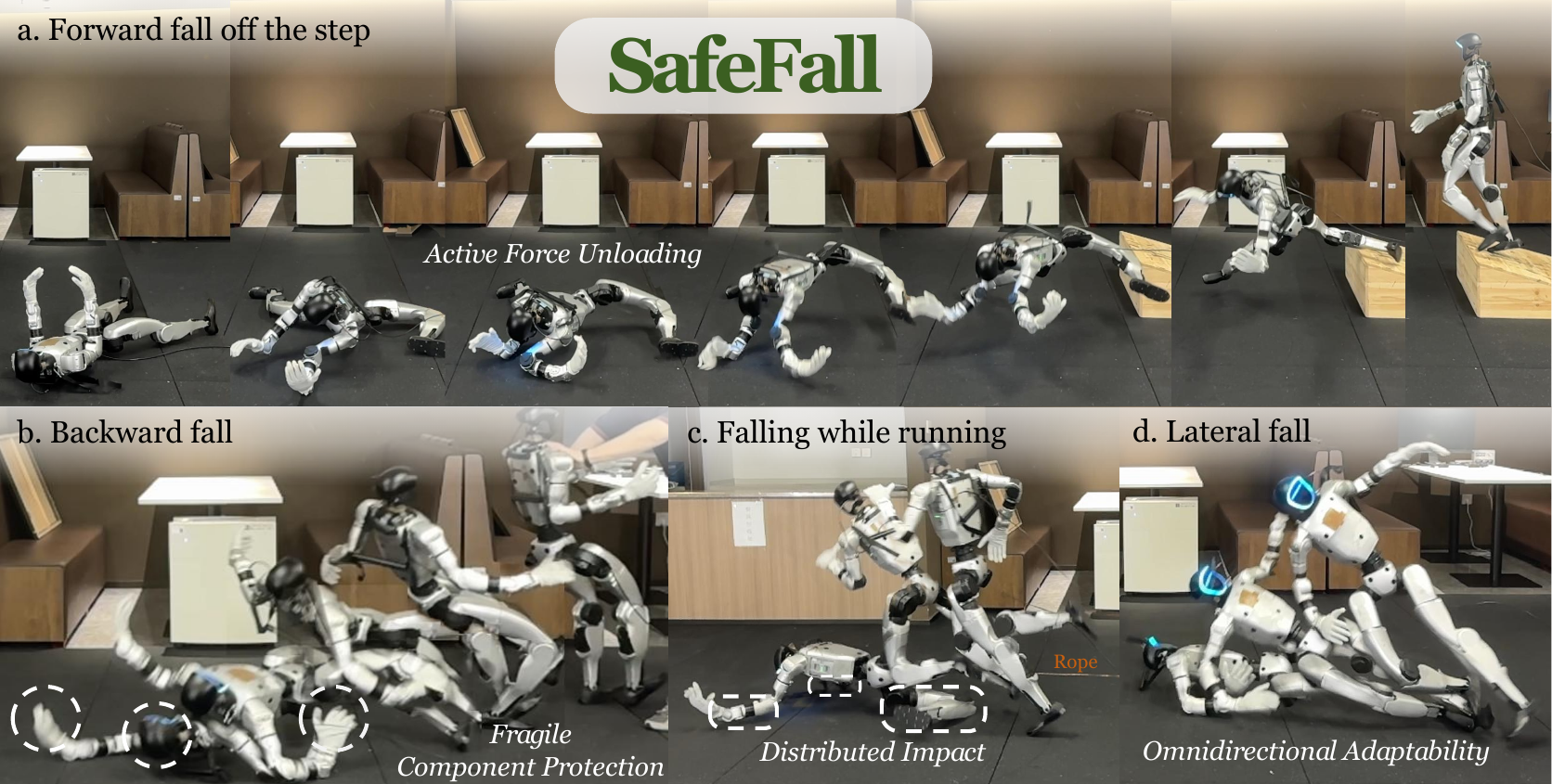}
\captionof{figure}{\method is the first method that protects full-scale humanoid robots from fall damages in the real world.
The proposed \method policy enables the humanoid to mitigate impact across a variety of scenarios, including (a) falling forward off the step, (b) falling backward, (c) rope-induced falls while running in 3 m/s, and (d) lateral falls. By prolonging the impact duration, distributing contact forces, and protecting fragile components, the policy achieves adaptive protection against complex, omnidirectional crashes.
}
\label{fig:teaser}
\end{strip}

\thispagestyle{empty}
\pagestyle{empty}


\newcommand{\pospart}[1]{\left[ #1 \right]_+}

\begin{abstract}

Bipedal locomotion makes humanoid robots inherently prone to falls, causing catastrophic damage to the expensive sensors, actuators, and structural components of full-scale robots. To address this critical barrier to real-world deployment, we present \method, a framework that learns to predict imminent, unavoidable falls and execute protective maneuvers to minimize hardware damage. \method is designed to operate seamlessly alongside existing nominal controller, ensuring no interference during normal operation. It combines two synergistic components: a lightweight, GRU-based fall predictor that continuously monitors the robot's states, and a reinforcement learning policy for damage mitigation. The protective policy remains dormant until the predictor identifies a fall as unavoidable, at which point it activates to take control and execute a damage-minimizing response. 
This policy is trained with a novel, damage-aware reward function that goes beyond minimizing external impacts. It explicitly models component heterogeneity to shield vulnerable areas like the head, while simultaneously constraining internal joint wrenches to prevent mechanical overload and electrical saturation of joint actuators.
Validated on a full-scale Unitree G1 humanoid, \method demonstrated significant performance improvements over unprotected falls. It reduced peak contact forces by 68.3\%, peak joint torques by 78.4\%, and eliminated 99.3\% of collisions with vulnerable components. By enabling humanoids to fail safely, \method provides a crucial safety net that allows for more aggressive experiments and accelerates the deployment of these robots in complex, real-world environments.
\end{abstract}

\section{Introduction}

The deployment of humanoid robots in real-world environments promises transformative applications in manufacturing, healthcare, and disaster response. However, their bipedal morphology introduces fundamental stability challenges that current control methods cannot fully address. Even state-of-the-art locomotion controllers remain vulnerable to falls caused by unexpected perturbations, perception delays, uneven terrain, and persistent sim-to-real gap in contact dynamics. For full-scale humanoid platforms weighing 30-80 kg, these falls pose a critical barrier to deployment: a single uncontrolled fall can destroy expensive sensors, damage actuators beyond repair, and render the entire system inoperable. While humans instinctively protect vital organs during falls through learned reflexive behaviors, humanoid robots currently lack analogous protective mechanisms. A robust fall protection system that operates transparently alongside nominal controllers would not only prevent catastrophic damage but also enable researchers to explore more aggressive control strategies without fear of hardware failure.

Prior work on fall mitigation primarily focused on small-scale robots or simulated environments, neither of which translates to modern humanoid systems. Small robots can employ strategies like hand cushioning \cite{yun2014tripod,ha2015multiple,xu2025unifiedhumanoidfallsafetypolicy} due to their favorable strength-to-weight ratios and simplified mechanical designs. However, full-scale humanoids present unique challenges: their dexterous manipulators contain delicate mechanisms that cannot withstand impact forces, their perceptual sensors (e.g. LiDAR, cameras) represent critical vulnerabilities worth thousands of dollars, and their higher kinetic energy during falls demands fundamentally different mitigation strategies. Moreover, existing approaches typically assume falls from static, upright positions, which is an unrealistic constraint given that real-world falls emerge from diverse failure modes during dynamic task execution.

Developing effective fall mitigation for humanoid robots presents three interconnected challenges. First, the strategy must exhibit structural awareness, which involves understanding which body components can safely absorb impact (e.g. reinforced torso shells) versus those requiring protection (e.g. sensor arrays, hand mechanisms). Second, the approach must handle distributional complexity, as real falls originate from varied initial conditions including different velocities, orientations, and center-of-mass trajectories that emerge from nominal policy failures. Third, training safe falling behaviors through reinforcement learning (RL) faces a fundamental temporal credit assignment problem: critical safety signals like peak impact forces occur only during brief collision events, while the majority of the falling trajectory provides sparse reward signals, making it difficult to learn which pre-impact actions minimize damage.

We present \method, a generalizable framework that enables humanoid robots to predict and respond to falls with damage-minimizing behaviors, seamlessly integrating with existing nominal controllers by monitoring continuously but intervening only when a fall is unavoidable. Our approach draws inspiration from human reflexive responses, extending impact duration through controlled rolling, protecting vulnerable areas, and maintaining post-impact stability. The framework comprises two synergistic components: (1) a GRU-based lightweight fall predictor trained on diverse failure scenarios that identifies irrecoverable states with minimal false positives, and (2) an RL policy that executes protective maneuvers prioritizing structural integrity. Critically, we introduce a novel training methodology that addresses the temporal credit assignment challenge by focusing episode boundaries on the fall-to-impact interval and incorporating damage models for specific components derived from actual robot specifications.

We validate \method through extensive experiments on a Unitree G1 platform, demonstrating significant reductions in peak joint torques $78.4\%$, contact forces $68.3\%$, and complete prevention of sensor damage across diverse fall scenarios. Real-world experiments confirm that our approach seamlessly integrates with existing nominal controllers while activating protective behaviors only when necessary, maintaining a false positive rate below $0.1\%$ and onboard inference time below $0.5$ms. This non-invasive design may fundamentally change the risk profile of humanoid robotics research, enabling more aggressive experimentation without catastrophic consequences.

Our contributions are threefold:
\begin{enumerate}
\item We present the first comprehensive fall-mitigation framework validated on full-scale humanoid robots in real-world conditions, incorporating platform-specific damage models and structural awareness into the learning process.
\item We develop a novel training pipeline that transforms nominal policy failures into a representative fall distribution, enabling robust fall prediction and mitigation across diverse operational scenarios.
\item We demonstrate that safe falling behaviors can be learned through RL despite sparse reward signals by introducing techniques to address the temporal credit assignment problem specific to metrics for impact events.
\end{enumerate}

\section{RELATED WORKS}

\subsection{Humanoid Robots Safe-falling}

Damage mitigation strategies for full-size humanoid robots have received comparatively little attention \cite{subburaman2023survey}, with most prior work conducted either in simulation or on small-scale platforms.
Early studies predominantly relied on model‑based motion planning to minimize impact velocity, reduce collision forces, and dissipate energy during falls \cite{ha2015multiple,wang2018unified,rossini2019optimal,Subburaman2018energyshaping,subburaman2018online,wang2017real}. Other works focused on triggering landing motions that direct impact toward less vulnerable body regions \cite{fujiwara2002ukemi,2011backpack}. While effective in structured settings, these methods generally depend on simplified dynamic models and accurate physical parameters, limiting their robustness in unstructured or uncertain environments.

More recently, RL has been applied to safe‑fall control, enabling policies that optimize for criteria such as minimizing impact impulses and protecting hardware components. 
Human fall demonstrations provide instructive insights, but the fundamental morphological and dynamical discrepancies between humans and robots limit the direct transferability of these strategies\cite{ruiz2009learning,yun2014tripod,kumar2017learning,buzzetti2024reinforcement}.
While concurrent work \cite{strauch2025robotcrashcourselearning} considers component sensitivity on a small-scale platform, it critically omits damage to actuators like joint motor. These approaches have demonstrated the potential for adaptive behavior; however, most have been tested on simplified robot morphologies or constrained to specific falling directions, reducing their applicability to full‑scale humanoids in diverse real‑world scenarios.
The main reasons are twofold.
First, they often learn a limited number of fixed optimal motion sequences on a restricted set of initial configurations. Such rigid strategies struggle to generalize across the wide variety of fall situations that may occur, particularly when transitioning abruptly from nominal control policies.
Second, these approaches typically neglect the differences in load bearing capacity, structural fragility, and material strength across different robot components.
In contrast, our proposed \method is explicitly designed to address these gaps. It accommodates a broad range of initial configurations, including divergent trajectories from failed nominal policies. Furthermore, we adopt joint-level physical constraints based on actuator mechanical and electrical limits. By integrating these robot-specific constraints into the learning process, achieving more robust and adaptable fall damage mitigation.

\subsection{Fall Prediction}

Humanoid stability and fall dynamics are critical issues in legged robotics, as they directly influence safety, hardware longevity, and overall performance. Advances in this domain have enabled timely fall prediction, which is essential for mitigating damage and facilitating rapid recovery.
Threshold-based approaches were among the earliest solutions, comparing sensor readings such as inertial measurement unit (IMU) data~\cite{5509550,8585406} to preset limits. While simple to implement, they rely on manual calibration and capture only narrow aspects of balance.
Model-based methods improved generality through principled criteria such as the Zero Moment Point (ZMP)~\cite{Ogata2008realtime}, which require accurate dynamic models but can be sensitive to parameter errors and unmodeled disturbances.
Learning-based strategies\cite{liu2020bidirectional,mungai2024fall,prabhakaran2025standing} leverage rich, multimodal inputs from IMUs, joint encoders, and vision sensors. Supporting this trend, Urbann et al.\cite{urbann2025dataset} released a large-scale dataset to accelerate data-driven fall prediction

Several existing works detect imminent instability and then trigger independent recovery controllers~\cite{mungai2024fall,prabhakaran2025standing}, rarely addressing early prediction of unavoidable falls or tightly coupling prediction with control. Our approach distinguishes itself from existing methods by integrating prediction and fall control, enabling early recognition of unavoidable falls. This integration ensures seamless coordination between detection and response, enabling faster and more effective fall damage mitigation, while the automatic labeling pipeline to avoid manual annotation cost.

\section{METHOD}

\begin{figure*}[t]
\centering
\includegraphics[width=1\textwidth]{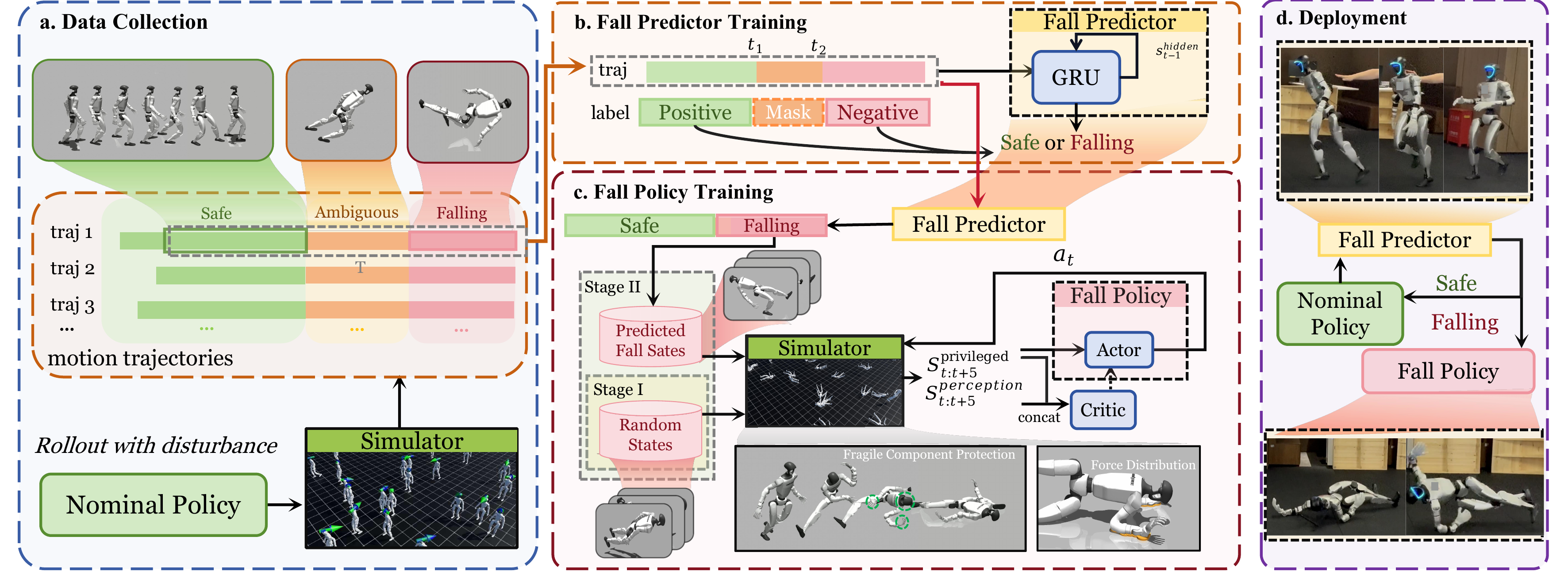}
\caption{Overview of \method. (a) We first collect diverse falling trajectories by rolling out the nominal locomotion policy under disturbances; (b) The trajectories are temporally segmented into safe, ambiguous, and falling phases to train a GRU-based fall predictor that identifies early signs of instability; (c) The \method\ policy is trained using a series of damage‑mitigation rewards to safeguard joint motors and other fragile components. To enhance robustness, training proceeds from random initial states to progressively realistic and challenging falling states; (d) During deployment, the fall predictor operates alongside the nominal policy, adaptively switching control to the fall policy upon detecting an impending fall for effective protection.
}
\label{fig:overview}
\end{figure*}



We present \method, a framework that addresses the dual challenge of predicting and mitigating fall damage in humanoid robots. Our approach integrates two synergistic components: a lightweight fall predictor that identifies irrecoverable states with minimal computational overhead, and an RL-based protection policy that executes behaviors minimizing damage upon fall detection. \cref{fig:overview} illustrates the system architecture and training pipeline.

\subsection{Preliminaries}

\textbf{Partially Observable Fall Mitigation}
We formulate humanoid fall mitigation as a Partially Observable Markov Decision Process $\langle\mathcal{S},\mathcal{A},T,\mathcal{O},R,\gamma\rangle$, where the challenge lies in learning protective behaviors from incomplete sensory information during the  
short critical 
window between fall onset and ground impact.

The observation space $\mathcal{O}$ contains only onboard sensor measurements: IMU readings and joint encoder measurements .  
Crucially absent are global root position and velocity, force measurements (both internal and external), which is information readily available in simulation but inaccessible during deployment.
The action space $\mathcal{A} \subset \mathbb{R}^{29}$ specifies target joint positions for the robot's PD controllers, bounded by the robot's joint position and velocity limits. $T(s^{\prime}|s,a)$ describes state transitions, and $\gamma \in (0, 1]$ is the discount factor. The goal is to learn a policy $\pi(a|o)$ that maximizes expected discounted returns while operating only on partial observations $o \in \mathcal{O}$.

The reward function $R(s,a)$ must balance competing objectives: minimizing peak impact forces while maintaining feasible joint configurations, protecting vulnerable components while utilizing robust body regions for energy absorption, and achieving stable behaviors during impact while avoiding unnecessary motion. We detail this multi-objective formulation in \cref{sec:fall_policy}.

\begin{figure}[t]
\centering
\includegraphics[width=0.45\textwidth]{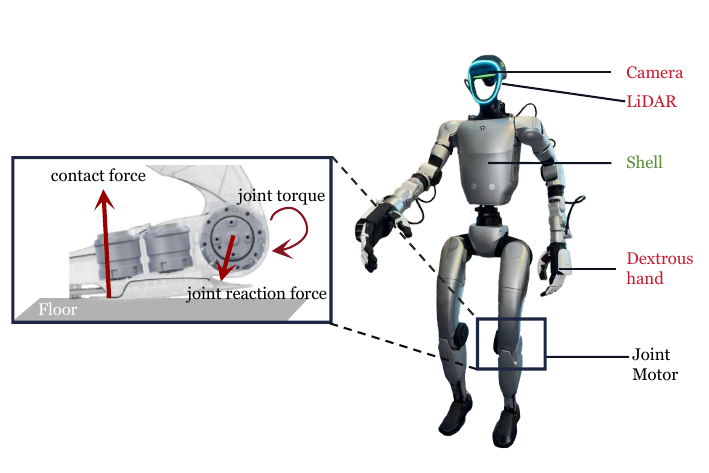}
\caption{Illustration of critical components in a humanoid robot. The perception modules and dexterous manipulators are susceptible to physical impact. In contrast, the arms, legs, and torso shell act as protective structures, attenuating external impacts, preserving structural integrity, and shielding the embedded electronics. On the left, we visualize the difference between contact force, joint torque, and joint reaction force.}
\label{fig:robot_parts}
\end{figure}

\begin{table*}[htbp]
\centering
\caption{Technical Causes of Humanoid Falls and Our Simulation Strategy to Induce a Fall}
\rowcolors{2}{gray!10}{white}
\begin{tabular}{p{2.4cm} p{5 cm} p{8.4cm}}
\toprule
\textbf{Failure Factors} 
& \textbf{Description}
& \textbf{Simulation Method to Induce a Fall} \\
\midrule
Sensor Noise & Noise or bias in sensors causing state estimation errors. & Add uniform noise to the policy observations with a magnitude $2$-$10 \times$ greater than the training configuration. \\
External Force & Unexpected forces disturb momentum and balance. & Apply a velocity perturbation to the torso, where the perturbation in the forward/backward direction (x‑axis) is uniformly sampled from $[-2,2]$~m/s, and the lateral direction (y‑axis) from $[-1, 1]$~m/s.\\
Foot Slip & Low friction leads to support foot slippage. &  Apply a 1 m/s horizontal velocity perturbation in a random direction to the stance foot.\\
Foot Trip & Swing foot hits obstacles or irregular terrain. & Use unseen terrain heightfields and obstacles with heights ranging from $[0, 15]$~cm. \\
System Delay & Latency in sensing, control, or actuation degrades stability. & Introduce a random delay uniformly sampled from $[0, 200]$~ms in the observation-to-action loop.\\
Dynamic Mismatch & Robot dynamics differ from control or training model. & Randomize physical parameters by sampling joint stiffness and damping from $[0.2,3]\times$ nominal values and applying a 0.1~m horizontal CoM offset to a random direction\\
\bottomrule
\end{tabular}
\label{tab:failure_modes}
\end{table*}

\subsection{Data Collection}
\label{sec:data collection}

Training robust fall prediction requires diverse failure trajectories that capture the transition from stable operation to irrecoverable falls. Since falls vary significantly across different robot tasks, we focus on omnidirectional locomotion as our primary domain, using a trained velocity-tracking controller as the nominal policy from which we induce failure behaviors.

We identify 6 primary failure factors from hardware experiments (\cref{tab:failure_modes}) and systematically reproduce them in simulation. These factors rarely occur in isolation since real falls typically result from multiple simultaneous failures, such as sensor drift combined with external forces or foot slippage during control delays.

To generate realistic failure scenarios, we apply targeted perturbations during locomotion (\cref{tab:failure_modes}), with each trajectory combining 1-3 factors based on empirically observed co-occurrence frequencies. This protocol yields 81,920 trajectories, each containing both recovery attempts from unstable postures and eventual falls. We allocate 65,536 sequences for training and 16,384 for validation. This dataset serves the training of both the fall predictor (\cref{sec:fall_predictor}) and the fall damage mitigation policy (\cref{sec:fall_policy}). 

\subsection{Fall Predictor}
\label{sec:fall_predictor}

We implement the fall predictor as a lightweight GRU network \cite{chung2014empirical} that processes proprioceptive observations to classify states as safe or falling. The single-layer architecture with 64 hidden units balances detection accuracy with the computational constraints of real-time deployment, achieving inference time under $0.5$ ms.

The predictor operates on the state vector $s_t = \{r_t, \omega_t, q_t, \dot{q}_t\}$, where $r_t$ denotes pelvis roll and pitch angles in the world frame, $\omega_t$ represents base angular velocity, and $q_t, \dot{q}_t$ are the angular positions and velocities of each joint relative to the default standing pose. These quantities are directly measurable through onboard IMU and joint encoders, eliminating dependency on external sensing or state estimation. The GRU processes states sequentially, maintaining temporal context through its hidden state, while a linear output layer produces binary fall/no-fall predictions.

Training the predictor faces a key challenge: determining when a robot transitions from recoverable imbalance to inevitable falling. This transition is gradual, not instantaneous, and depends on the robot's complex dynamics and initial conditions. We resolve this through conservative temporal segmentation that creates clear training labels while preserving prediction lead time.

For each trajectory of length $T$ where $T$ is the time of ground impact, we define boundaries $t_1 = \frac{2T}{3}$ and $t_2 = T - 100~\text{ms}$, creating three segments: safe states $\mathcal{D}_1$ for $t \leq t_1$, ambiguous states $\mathcal{D}_u$ for $t_1 < t \leq t_2$, and falling states $\mathcal{D}_2$ for $t > t_2$. The safe segment intentionally includes recoverable instabilities to prevent false alarms during aggressive but controlled maneuvers. The 100 ms falling window represents the minimum reliable detection horizon before impact.

We train the fall predictor using a negative log-likelihood loss with the ambiguous segment $\mathcal{D}_u$ masked out from gradient computation, preventing the model from learning uncertain labels.

\subsection{\method Policy}\label{sec:fall_policy}

Upon detecting an irrecoverable fall, the system transitions from the nominal controller to a specialized mitigation policy trained to minimize impact damage. We learn this policy using PPO~\cite{schulman2017proximal} with a reward function that explicitly models component vulnerability alongside motor mechanical and electrical constraints. 
To address the temporal credit assignment problem inherent in sparse impact signals, we restrict training to a fixed, short episode length the detection of an unavoidable fall to the completion of ground impact.

\textbf{Two-Stage Curriculum Learning} 
We employ a curriculum approach to balance computational efficiency with physical realism. Stage I uses simplified collision geometry and basic initial configuration to rapidly explore fall strategies, while stage II refines the policy using full collision models and initializes training from realistic falling states.
To ensure robust fall mitigation across diverse failure modes, we employ different initialization strategy in two stages. In Stage I, we generate stochastic falling configurations by placing the robot in random poses slightly deviating from its default pose and random orientations slightly above the ground. In Stage II, we sample states from our collected dataset that the fall predictor classifies as unsafe. We introduce this second sampling method only in Stage II to progressively increase training difficulty. In both stages, kinematically invalid configurations including ground penetration and self-collision are filtered.

\textbf{Damage-Aware Reward} The reward function balances two objectives: minimizing collision impact loads that could damage hardware, and maintaining stable, efficient motions during the fall. We formulate the total reward as 
\begin{equation}
r_\text{total}=r_\text{impact}+r_\text{regulation}
\end{equation}
where $r_{\text{impact}}$ captures physical damage risks and $r_{\text{regulation}}$ ensures motion quality.

We adopt a decoupled approach to separately measuring collisions between adjacent links from those between non-adjacent links and with the ground. When employing full collision models, geometric limitations of the collision meshes often cause minor overlaps between adjacent links that are precisely articulated in the real world. These overlaps lead the physics solver to compute enormous, non-physical anomalous contact forces, distorting the collision reward signal.
We posit that the essential nature of the physical interaction between adjacent links is a manifestation of joint internal constrains.
Consequently, we exclude contact forces between adjacent links and instead rely on the joint reaction force to quantify these internal constrains. These reaction forces properly model the physical response when adjacent links undergo \textit{structural collisions} or bear destructive stress. Moreover, joint torque penalty serve to constrain motor efforts, preventing them from exceeding the electrical performance of actuators.

Consequently, we model impact safety through three complementary terms:
\begin{equation}
r_\text{impact}=w_c\cdot r_\text{contact}+w_j\cdot r_\text{joint} + w_e \cdot r_\text{torque}.
\end{equation}

The primary contact force penalty accounts for component heterogeneity by assigning sensitivity weights $w_{s,i} \in {1000, 1, 0.5}$ to high (e.g. head, hands), medium (e.g. shanks, shoulders), and low (e.g. torso, thighs, elbows) vulnerability regions based on replacement costs and functional criticality:
\begin{align}
r_{\text{contact}} &=\frac{1}{N} \sum_{i=1}^{L} \left\| \mathbb{I}\{c_i\} \; w_{s,i} \;
    \pospart{f_{\text{contact},i} - m_i g} \nonumber \right\|_2 \\
    & \quad + \alpha \cdot
     \max_{i \in \{1,\dots,L\}} \left\| \mathbb{I}\{c_i\} \; w_{s,i} \;
    \pospart{f_{\text{contact},i} - m_i g} \right\|_2
\end{align}
where $\mathbb{I}\{c_i\} = 1$ if link $i$ is in contact and 0 otherwise, $f_{\text{contact},i}$ is the contact force magnitude on link $i$, $m_i g$ represents gravitational loading (subtracted to isolate dynamic contact force), $\pospart{x} = \max(x, 0)$ ensures non-negativity, $N = \sum_{i=1}^{L} \mathbb{I}{c_i}$ counts active contacts to encourage force distribution, and $\alpha = 0.3$ balances average and peak forces. As mentioned above, anomalous contact forces arising from spurious collisions between adjacent links due to imprecise collision models are also filtered. 

The joint reaction force penalty protects mechanical connections from excessive loads:

\begin{equation}
    r_{\text{joint}} = 
    \sum_{i=1}^{J}
    \big\| f_{\text{joint},i} - f^{\text{thresh}}_{\text{joint},i} \big\|_2
\end{equation}
where $f_{\text{joint},i}$ represents joint reaction forces computed by the physics solver to maintain joint kinematics during impacts. These forces reflect mechanical loads transmitted through the kinematic chain, as they arise when joints experience reaction forces to maintain their prescribed degrees of freedom during ground impact propagation. The threshold $f^{\text{thresh}}_{i}$ corresponds to each joint's mechanical load capacity determined from hardware specifications.

The torque penalty prevents actuator saturation and mechanical stress:

\begin{equation}
r_{\text{torque}} =
\sum_{i=1}^{J} 
\left\| \pospart{\frac{\tau_i}{\bar{\tau}_{i}} - 1} \right\|_2 ,
\end{equation}
where $\tau_i$ represents external torques on joint $i$ from constraint reactions and external loads (excluding motor output), and $\bar{\tau}_{i}$ is the actuator's maximum rated torque. This ensures the policy maintains control authority during impacts while avoiding damaging torque spikes.

Additional regularization terms $r_{\text{regulation}}$ penalize excessive joint positions, velocities, accelerations, and action rates. 
Beyond promoting smooth motions and suppressing oscillatory behaviors that frequently emerge during ground contact transitions, these regularization terms also play a critical role in reducing collision related artifacts by discouraging joint limit strikes and abrupt accelerations. Collectively, these terms constrain undesirable collision behaviors, as well as improving both simulation stability and hardware transferability.

\textbf{Asymmetric Actor-Critic} The actor network observes only deployable sensor measurements: pelvis orientation $(r, p)$, joint states $(q, \dot{q})$, previous actions $a_{t-1}$, angular velocity $\omega$, and projected gravity $\mathbf{g}_b$, stacked over 5 timesteps for temporal context. The critic additionally accesses privileged simulation state including global root positions, velocities and center-of-mass, enabling more effective value estimation during training. This asymmetric design \cite{christiano2016transfer} maintains deployment feasibility while exploiting rich supervision.

\textbf{Domain Randomization}
To facilitate robust sim-to-real transfer, we employ domain randomization during training, perturbing both observations and physical parameters to improve the policy’s resilience to real-world dynamics. We detail the noise and randomization parameters in \cref{tab:domain_rand_noise}.

\begin{table}[h]
\caption{Domain randomization and observation noise configurations used during training.}
\label{tab:domain_rand_noise}
\centering
\begin{tabular}{p{3cm} p{3cm}}
\hline
\textbf{Term} & \textbf{Value} \\
\hline
\multicolumn{2}{c}{\textbf{Domain Randomization}} \\
\hline
Friction & $\mathcal{U}(0.3, 1.0)$ \\
Restitution & $\mathcal{U}(0.0, 0.5)$ \\
Base mass offset (kg) & $\mathcal{U}(-1.0, 3.0) $ \\
Base CoM offset (m) & $x,y \sim \mathcal{U}(-0.05,0.05),\; \newline z \sim \mathcal{U}(-0.01,0.01)$ \\
Joint stiffness scale & $\log \mathcal{U}(0.7, 1.5)$ \\
Joint damping scale & $\log \mathcal{U}(0.5, 3.0)$ \\
Joint position limits & $\mathcal{N}(0, 0.02)$ \\
\hline
\multicolumn{2}{c}{\textbf{Observation Noise}} \\
\hline
Root quat (--) & $\mathcal{U}(-0.05, 0.05)$ \\
Joint pos (rad) & $\mathcal{U}(-0.01, 0.01)$ \\
Joint vel (rad/s) & $\mathcal{U}(-1.5, 1.5)$ \\
Base ang vel (rad/s) & $\mathcal{U}(-0.2, 0.2)$ \\
Projected gravity (--) & $\mathcal{U}(-0.05, 0.05)$ \\
\hline
\end{tabular}
\end{table}

\subsection{Training and Implementation Details}
Our experiments are conducted on a Unitree G1 humanoid robot~\cite{unitree2024g1} with 29 degrees-of-freedom (DoF). 

We train the fall prediction model using the Adam optimizer with a learning rate of $10^{-3}$ and weight decay of $10^{-4}$ for 5 epochs. We use a batch size of 4096. The training takes only about 5 minutes on a single RTX 4090 GPU.  

For the \method policy, we use a fixed episode length of 40. Both the Actor and the Critic are implemented as a 3-layer MLP. The policy is trained using the Adam optimizer with an initial learning rate of $10^{-3}$ along with an adaptive learning rate scheduling scheme. 
Training \method takes 128 GPU hours for Stage I and 152 GPU hours for Stage II, amounting to a total of 280 GPU hours.
A proportional–derivative (PD) controller operates at 200Hz in both simulation and on the real-world humanoid platform. In real-world deployment, both the predictor and the \method policy runs at 50Hz.

\begin{figure}[t]
\centering
\includegraphics[width=0.45\textwidth]{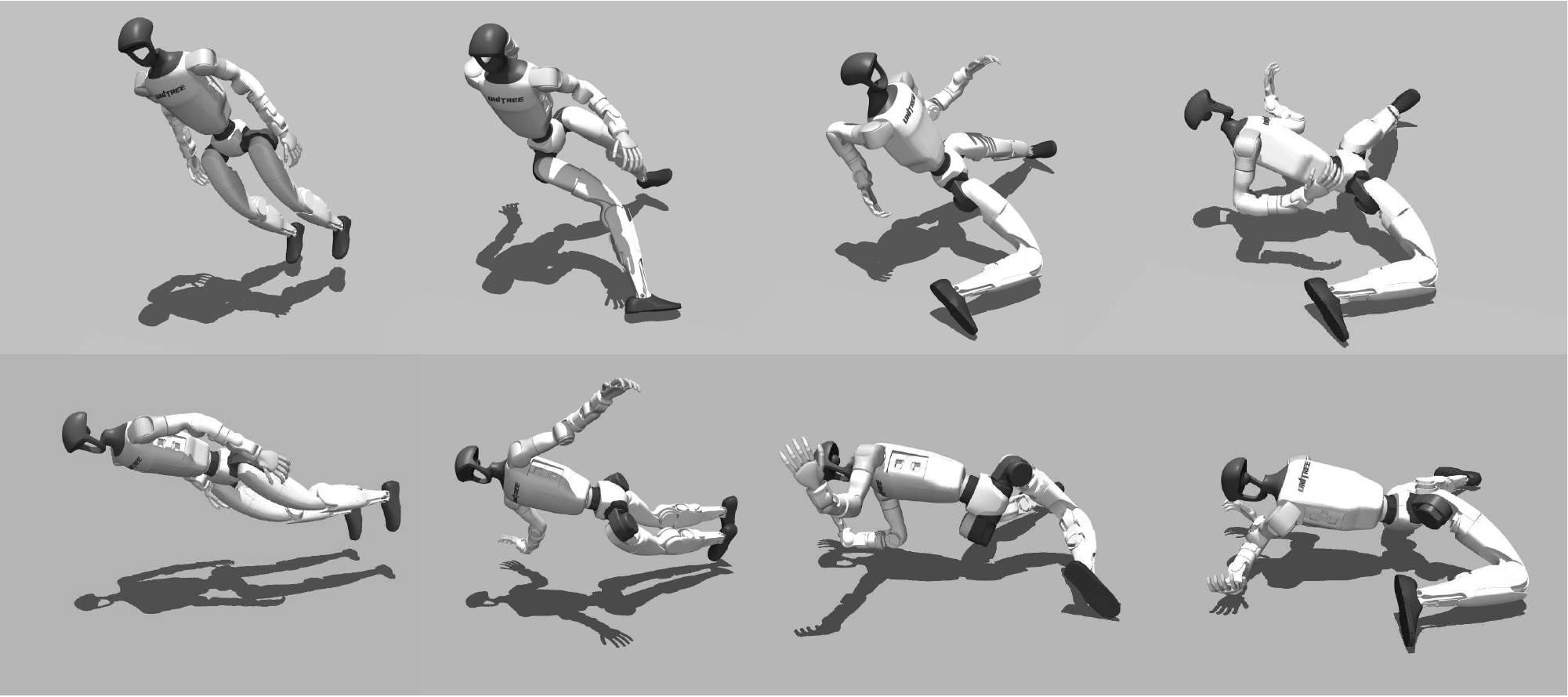}
\caption{
Robot falling in different directions.
\textbf{Top: Lateral fall.} The robot rotates its upper body to land on its high priority rigid torso, orienting its camera-equipped head upward while using its arms for auxiliary cushioning. \textbf{Bottom: Forward fall.} The robot extends its arms to brace for impact, absorbing energy and protecting its head from direct impact. 
}
\label{fig:endpose}
\end{figure}

\section{Experiments}

We evaluate our fall protection system through three key research questions:

\textbf{RQ1}: Does the fall predictor achieve sufficient specificity to avoid false triggers while providing adequate warning time?

\textbf{RQ2}: How effectively does our damage-aware policy reduce impact forces compared to baseline strategies?

\textbf{RQ3}: Can the learned mitigation policy generalize to different locomotion controllers?




\subsection{Evaluation of Fall Predictor}
We evaluate predictor reliability using False Alarm Rate (FAR), the fraction of stable states misclassified as falling, and Lead Time (LT), the interval between first detection and ground impact. \Cref{tab:predictor} compares our GRU-based predictor against a 5-frame sliding-window MLP baseline and GRU-based predictors with different training data segmentation boundaries $t_2$.

The GRU-based predictor with $t_2=T-100$ ms achieves a favorable balance between detection accuracy and response time, maintaining FAR well below 0.1\% while providing 410ms lead time. The results suggest that temporal modeling provides meaningful advantages in fall detection scenarios. While the MLP's sliding-window approach captures local temporal patterns through frame concatenation, the GRU's recurrent architecture offers better discrimination between transient instabilities during aggressive maneuvers and genuine balance loss events. 

Interestingly, a larger $t_2$ value significantly increases the lead time at minimal FAR cost (0.06\% to 0.16\%), suggesting that the predictor can learn discriminative features even with relaxed thresholds. This enables application specific tuning, with conservative settings for environments where false triggers are costly, versus aggressive detection where hardware protection is paramount.

Furthermore,  masking out the uncertain region between $t_1$ and $t_2$ during training significantly improves the FAR and the lead time across both architectures, suggesting the value of this technique in addition to the chosen temporal modeling approach.

\begin{table}[htbp]
\centering
\caption{Ablation of fall predictor design choices}
\renewcommand{\arraystretch}{1.2}
\begin{tabular}{l|cc}
\hline
\textbf{Method}  $\uparrow$& \textbf{FAR} $\downarrow$ & \textbf{LT(s)} $\uparrow$ \\
\hline

\hline
 $without~ masked ~\mathcal{D}_u$\\
Baseline ($t_1=t_2=T-0.2s$)  & $  0.71\%$ & $ 0.24 $ \\
Fall Predictor ($t_1=t_2=T-0.2s$)  & $  0.35\%$ & $ 0.18 $ \\
Fall Predictor ($t_1=t_2=\frac{2T}{3}$)  & $  4.1\%$ & $1.22$ \\
\hline
$masked ~\mathcal{D}_u$ \\
Baseline ($t_1=\frac{2T}{3},t_2=T-0.2s$)  & $ 0.09 \%$ & $ 0.44 $ \\
Fall Predictor ($t_1=\frac{2T}{3}, t_2=T-0.1s$)  & $ \textbf{0.04\%} $ & $ 0.29 $ \\
\underline{Fall Predictor ($t_1=\frac{2T}{3},t_2=T-0.2s$)}  & $\underline{0.06\%}$ & $ \underline{0.41} $ \\
Fall Predictor ($t_1=\frac{2T}{3},t_2=T-0.4s$)  & $ 0.16 \%$ & $ \textbf{0.70} $ \\
\hline
\end{tabular}
\label{tab:predictor}
\vspace{-6pt}
\end{table}


\subsection{Damage Mitigation}

We quantitatively assess the effectiveness of our approach in mitigating fall-induced damage in simulation. The adopted metrics are as follows, where each metric reports the peak magnitude observed during a single trial. To simplify notation, we omit the superscript $^{\text{max}}$ in the remainder of this section.

\begin{itemize}[] %
    \item \textbf{Max Joint Torque} $\tau$: Maximum joint torque over all joints. This metric captures the peak rotational loads experienced by the actuators
    \item \textbf{Max Joint Reaction Force} $f_\text{joint}$: Maximum joint reaction force magnitude over all joints. Joint reaction forces are the internal constraint forces required to maintain the kinematic connection between adjacent links, thereby reflecting the destructive stress acting on the joints.
    \item \textbf{Max Contact Force} $f_\text{contact}$: Peak contact force between body links and ground during impact.
    \item \textbf{Whole-Body Impulse} $J$: The maximum ground reaction impulse exerted on the robot within a single simulation time step. This metric captures the most severe instantaneous impact spike..
    \item \textbf{Illegal Contact} $N_\text{illegal}$: The number of collisions on protected regions (e.g., the head or dexterous hands).
    \item \textbf{Joint Limit Violations} $N_\text{limit}$: The maximum number of joints violating their joint position limits at any single time step.
    \item \textbf{Max Joint Torque Ratio} $R_\text{torque}$: The maximum ratio of the torque experienced by any articulated joint to its maximum designed torque capacity, evaluated over all joints.

\end{itemize}

We compare against three baselines: 
\begin{itemize}
    \item \textbf{Nominal Policy}: Continuing the nominal locomotion policy through impact.
    \item \textbf{Default Pose}: Maintaining upright standing posture with arms and legs held closely to the body.
    \item \textbf{Damping Mode}: Switching to damping mode ($K_p=10e^{-5}~(\approx 0), K_d=10$) upon fall detection, with constant target position and zero target velocity.
\end{itemize}

We evaluated \method on over 5,000 falling scenarios, initialized with diverse fall postures and velocities spanning $[0, 4]$~m/s. These scenarios were generated following the same initialization protocol from Stage II training, ensuring they represent realistic failure modes.
\cref{tab:fallpolicy} demonstrates substantial safety improvements in terms of all metrics. Compared to the nominal policy, our method reduces peak joint torque, joint reaction force, and contact force by 78.4\%, 66.8\%, and 68.3\%, respectively, indicating significantly lower structural damage risk. 
The relatively large standard deviation arises from the influence of the initial conditions (e.g. height, posture and velocity) on the resulting impact dynamics.
Protected region contacts drop from 56-99\% in baselines to less than 1\% with our approach, effectively safeguarding vulnerable components. The lowest cumulative impulse confirms smoother energy dissipation through controlled protective motions.
We note that the Max Joint Torque Ratio exceeds 1.0 (specifically 2.47) during impact. This remains within the physical safety margins of the actuators~\cite{siciliano2008springer}. Joint actuators are typically designed to withstand transient peak torques higher (often 3--5$\times$) than their rated continuous torque, particularly during short-duration events such as ground impact.

\begin{figure}[t]
\centering
\includegraphics[width=0.45\textwidth]{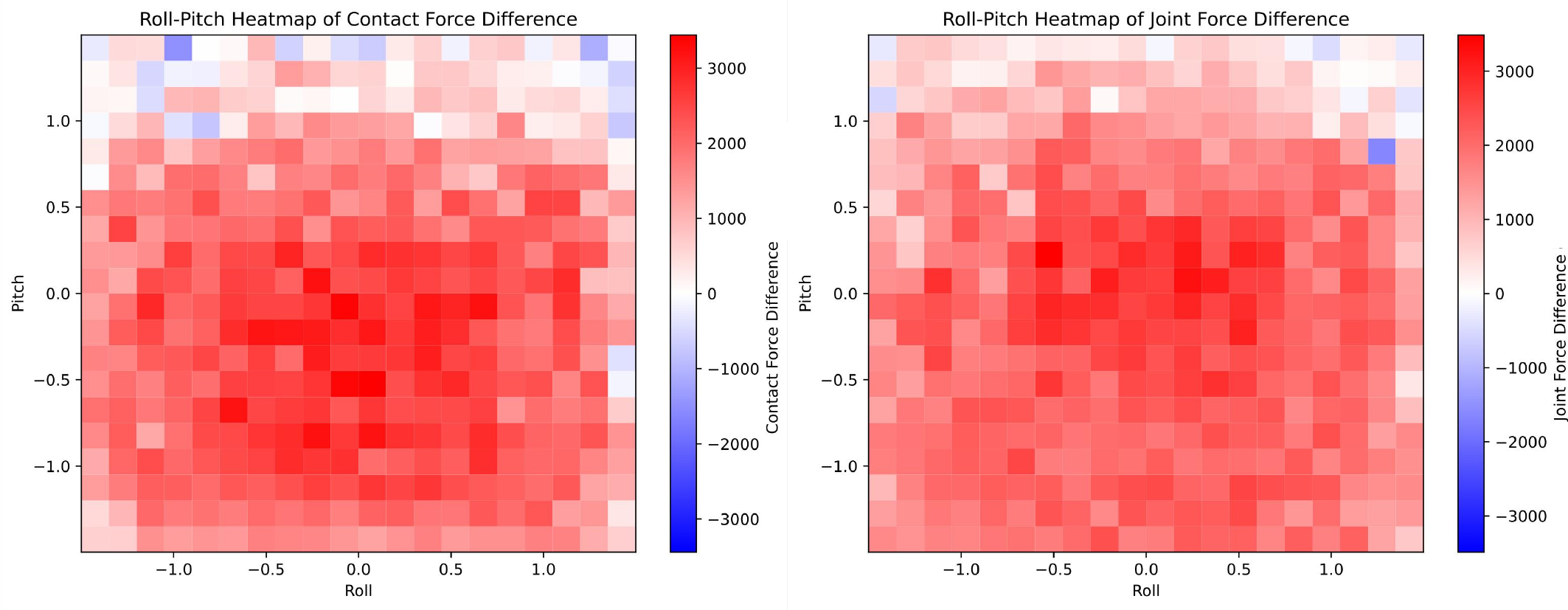}
\caption{
Improvement in maximum contact force (left) and max joint force (right) achieved by \method compared to the damping mode when the robot falls at various roll and pitch angles. \method achieves significant improvement in most fall scenarios, highlighting its robust performance in omnidirectional falls.
}
\vspace{-6pt}
\label{fig:improvement}
\end{figure}

\Cref{fig:improvement} analyzes directional robustness by evaluating falls across the full range of roll and pitch orientations.
Our method consistently outperforms the best baseline (damping mode) across all directions, with particularly large improvements for lateral and forward falls where uncontrolled impacts typically concentrate on vulnerable regions. This omnidirectional effectiveness answers RQ2 affirmatively.

Qualitative examples of the learned  damage-mitigation behaviors for forward and lateral falls are visualized in \cref{fig:endpose}. Notably, in most cases, the robot retracts its legs to actively lower its pelvis. This action ensures that the lower body contacts the ground first, preventing the upper body, particularly the head, from becoming the primary point of impact.
We also observed that the policy exploits the unique kinematics of the robot to execute non-anthropomorphic behaviors that are impossible for humans, such as 180 degree waist yaw rotation.
This highlights that the learned policy is driven by the intrinsic hardware characteristics of robot, rather than being constrained by an anthropomorphic perspective.

\begin{table*}[t]
\centering
\caption{Qualitative evaluations of \method reveal consistent and significant improvement over baseline methods}
\renewcommand{\arraystretch}{1.2}
\begin{tabular}{l|ccccccc}
\hline
\textbf{Method} 
& \textbf{$\tau$(N$\cdot$m)} $\downarrow$
& \textbf{$f_\text{joint}$(N)} $\downarrow$
& \textbf{$f_\text{contact}$(N)} $\downarrow$ 
& \textbf{$J$ (N$\cdot$s)} $\downarrow$ 
& \textbf{$N_\text{illegal}$} $\downarrow$
& \textbf{$N_\text{limit}$} $\downarrow$
& \textbf{$R_\text{torque}$} $\downarrow$
\\
\hline
Nominal Policy & $613 \pm 401$ & $4096 \pm 3058$ & $ 4036 \pm 2542$ & $426$ &$ 99\%$ & $3.2 $ & $12.45$\\
Default Pose & $ 402 \pm 384$ & $3064 \pm 2744$ & $ 3132 \pm 2383$ & 260.1 & $ 82 \%$ & $ 0.21$ & $ 6.44 $\\
Damping Mode & $304 \pm 241$ & $2844 \pm 2482$ & $ 3138 \pm 2575$ & 220.4 & $ 56\%$ & $ 0.96 $ & $5.73$\\
\hline
Ours (Stage I) & $295 \pm 240$ & $ 2360 \pm 1908$ & $2768 \pm 2113$ & $370$ & \textbf{$ 43\%$} & \textbf{$ 0.57$} & $7.09$\\
Ours (Stage I\&II)& $\mathbf{132 \pm 76}$ & $\mathbf{1361 \pm 1351}$ & \textbf{$ \mathbf{1279 \pm 1008}$} & $\mathbf{180.7}$ & $ \mathbf{0.7\%}$ & {$ \mathbf{0.003}$} & $\mathbf{2.47}$\\

\hline
\end{tabular}
\label{tab:fallpolicy}
\end{table*}


\subsection{Generalization}
To evaluate generalization across nominal policies (RQ3), we deploy our fall mitigation policy on a stylized locomotion controller~\cite{ma2025styleloco} with significantly different gait patterns from the training distribution. While the fall predictor requires retraining due to altered motion signatures, the mitigation policy transfers directly without fine-tuning. \cref{tab:styleloco} shows significant damage reduction (48.9\% max joint force decrease and 50.0\% max contact force decrease) despite the distribution shift, confirming that training on diverse fall trajectories enables robust generalization to unseen locomotion policies. 

\begin{table}[htbp]
\centering
\caption{\method generalizes to an unseen nominal policy}
\label{tab:styleloco}
\renewcommand{\arraystretch}{1.2}
\begin{tabular}{l|cc}
\hline
\textbf{Method}  
& \textbf{$f_\text{joint}^\text{max}$(N)} $\downarrow$
& \textbf{$f_\text{contact}^\text{max}$(N)} \\
\hline
Damping Mode  & $ 4780 $ & $ 5287 $ \\
Ours  & $ 2444 $ & $ 2644 $ \\
Improvement & $-48.9\%$ & $ -50.0\%$ \\
\hline
\end{tabular}
\end{table}

\begin{figure}[t]
\centering
\includegraphics[width=0.4\textwidth]{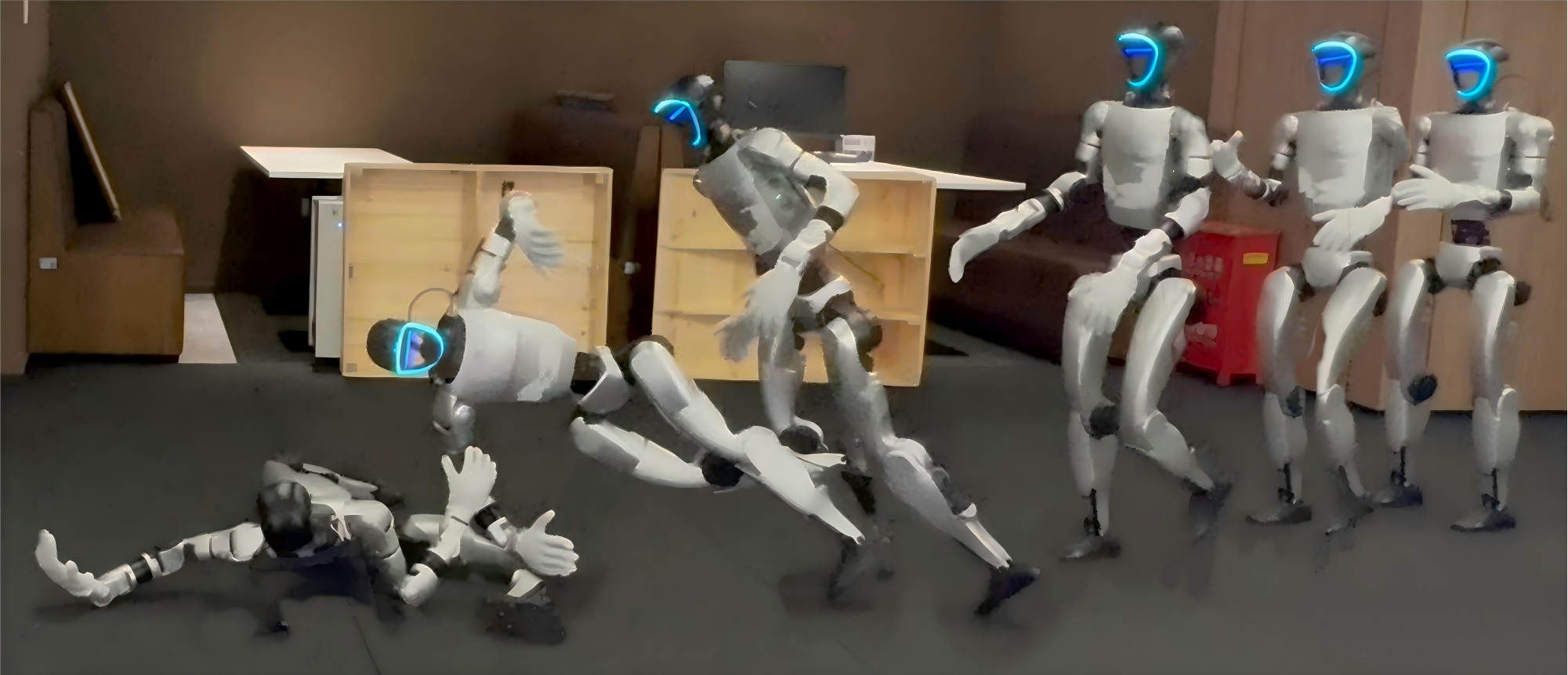}
\caption{\method mitigating damages from a fall. We observe clear protective motions in which the robot rotates and extends its arm to absorb energy and shield its head and hands from heavy impact.}
\label{fig:real}
\end{figure}

\subsection{Real-world Experiments}
We conducted real-world experiments on a Unitree G1 robot to validate \method's effectiveness on physical hardware through a comprehensive suite of perturbation experiments. This suite includes external pushes from different directions (forward, backward, and lateral) during walking, alongside challenging scenarios such as misstepping off a 30 cm platform and tripping during 3 m/s high-speed running.

\cref{fig:teaser,fig:real} and the accompanying video demonstrate that \method achieves clear fall damage mitigation behaviors across these crash scenarios.
Notably, the robot learned to utilize its elbows to brace against the ground, effectively distributing impact loads. Simultaneously, the policy exhibited delicate wrist reorientation maneuvers to actively avoid ground contact by flipping the hands upward. These learned behaviors effectively shielded critical components from direct impact.

To quantify impact reduction, we measured peak impulse over 100ms during the fall using a high-speed motion capture system. Our method achieved 286.1 N$\cdot$s peak impulse—a 22.1\% reduction compared to the damping baseline's 367.1 N$\cdot$s. This dramatic decrease matches our simulated experiments and validates our approach's protective capabilities. 

Additionally, the system demonstrated perfect specificity: the predictor exhibited zero false positives during controlled push experiments, correctly distinguishing recoverable disturbances from impacts that induce falls. Gentle pushes that allowed natural recovery did not trigger protection mechanism, while strong disturbance reliably activated the mitigation response.

\section{CONCLUSION AND LIMITATIONS}

We presented \method, a framework for fall prediction and damage mitigation in full-scale humanoid robots. Our system combines a lightweight GRU-based fall predictor achieving 0.06\% false alarm rates with an RL policy that reduces peak contact forces by 68.3\%, joint torques by 78.4\%, and joint reaction forces by 66.8\% compared to unprotected falls. The framework eliminates 99.3\% of collisions with vulnerable components such as sensors and dexterous hands.

Critically, \method operates alongside existing control systems with minimal computational overhead. During normal operation, only the predictor runs, continuously monitoring for irrecoverable states without interfering with the nominal controller. The protective policy activates only when a fall becomes inevitable, replacing the nominal controller to execute damage-minimizing behaviors. This design ensures zero performance degradation during standard operation while providing crucial protection when needed.

To our knowledge, this work represents the first fall protection framework validated on full-scale humanoid platforms in real-world experiments. The proposed framework may fundamentally change the risk profile of humanoid robotics research: researchers can experiment with more aggressive control strategies and push performance boundaries, knowing that inevitable failures will not result in catastrophic hardware damage. This safety net reduces both repair costs and development downtime, accelerating the iteration cycle for novel control approaches.

While we validated \method on the Unitree G1 platform performing locomotion tasks, the training methodology, including automated fall data generation and component-specific damage modeling, can be applied to other humanoid platforms and tasks. The framework requires only proprioceptive sensing and knowledge of component vulnerabilities, making it compatible with existing humanoid systems.

\textbf{Limitations and Future Work}
Our approach has two primary limitations. First, training convergence requires approximately 280 GPU hours due to the sparse reward structure inherent in impact-based optimization. The critical learning signals occur only during brief collision events, while the majority of the falling trajectory provides minimal gradient information. This computational cost affects only the initial training phase, not deployment. Moreover, the trained fall policy is generalizable to different nominal policies.

Second, the current system operates reliably only on flat or near-flat terrain. Stairs, ledges, and significantly uneven surfaces present fundamentally different fall dynamics that our policy cannot adequately address. These terrain features pose particular challenges as they often cause the most severe damage in real-world deployments. Extending \method to handle complex terrain requires visual perception to identify geometric hazards and adapt protective strategies accordingly. This represents an important direction for future research, as vision-based fall mitigation would enable safe operation in the diverse environments where humanoids are intended to operate.

Despite these limitations, \method provides a practical solution for reducing fall-related damage in humanoid robots, contributing to their deployment in real-world applications where falls are inevitable but must not be catastrophic.


\bibliography{reference}
\bibliographystyle{ieeetran}
\end{document}